\documentclass[11pt,a4paper]{article}
\usepackage[hyperref]{emnlp2018}
\usepackage{times}
\usepackage{latexsym}
\usepackage{subcaption}
\usepackage[utf8]{inputenc}
\usepackage{url}
\usepackage{multirow}
\usepackage{amsmath}
\usepackage{tikz}
\usepackage{todonotes}
\usepackage{tikz}
\usetikzlibrary{arrows, calc, fit}

\tikzset{
	font=\rmfamily \small,
	word/.style = {
		draw,
		semithick,
		rounded corners = 6pt,
		minimum size = .9cm,
		inner sep = .3cm
	},
	ig/.style = {
		word,
		thin,
		rounded corners = 4pt,
		densely dashed
	},
	root/.style = {
		word,
		dotted
	},
	igset/.style = {
		word,
		inner sep = .16cm
	},
	dep/.style 2 args = {
		->,
		> = stealth,
		semithick,
		rounded corners = 6pt,
	},
	deplink/.style 2 args = {
		-,
		semithick,
		rounded corners = 6pt,
	},
	ubd/.style 2 args = {
		dep,
		thin,
		densely dashdotted
	},
	ubdlink/.style 2 args = {
		deplink,
		thin,
		densely dashdotted
	},	
	textabove/.style = {
		font = \ttfamily \tiny,
		inner sep = .03cm,
		above
	},
	textbelow/.style = {
		font = \ttfamily \tiny,
		inner sep = .03cm,
		below
	},
	ldep/.style 2 args = {
		dep,
		to path = ($ ($ (\tikztostart)!0.2cm!(\tikztotarget) $)!0.45cm!90:(\tikztotarget) $)
		       -- ($ ($ (\tikztostart)!0.2cm!(\tikztotarget) $)!#1!90:(\tikztotarget) $)
		       -- node[textabove] {#2}
		          ($ ($ (\tikztotarget)!0.2cm!(\tikztostart) $)!#1!-90:(\tikztostart) $)
		       -- ($ ($ (\tikztotarget)!0.2cm!(\tikztostart) $)!-0.45cm!90:(\tikztostart) $)
	},
	rdep/.style 2 args = {
		dep,
		to path = ($ ($ (\tikztostart)!0.2cm!(\tikztotarget) $)!0.45cm!90:(\tikztotarget) $)
		       -- ($ ($ (\tikztostart)!0.2cm!(\tikztotarget) $)!#1!90:(\tikztotarget) $)
		       -- node[textbelow] {#2}
		          ($ ($ (\tikztotarget)!0.2cm!(\tikztostart) $)!#1!-90:(\tikztostart) $)
		       -- ($ ($ (\tikztotarget)!0.2cm!(\tikztostart) $)!-0.45cm!90:(\tikztostart) $)
	},
	lubd/.style 2 args = {
	  ubd,
		to path = ($ ($ (\tikztostart)!0.2cm!(\tikztotarget) $)!0.45cm!90:(\tikztotarget) $)
		       -- ($ ($ (\tikztostart)!0.2cm!(\tikztotarget) $)!#1!90:(\tikztotarget) $)
		       -- node[textabove] {#2}
		          ($ ($ (\tikztotarget)!0.2cm!(\tikztostart) $)!#1!-90:(\tikztostart) $)
		       -- ($ ($ (\tikztotarget)!0.2cm!(\tikztostart) $)!-0.45cm!90:(\tikztostart) $)
	},
	rubd/.style 2 args = {
		ubdlink,
		to path = ($ ($ (\tikztostart)!0.2cm!(\tikztotarget) $)!0.45cm!90:(\tikztotarget) $)
		       -- ($ ($ (\tikztostart)!0.2cm!(\tikztotarget) $)!#1!90:(\tikztotarget) $)
		       -- node[textbelow] {#2}
		          ($ ($ (\tikztotarget)!0.2cm!(\tikztostart) $)!#1!-90:(\tikztostart) $)
		       -- ($ ($ (\tikztotarget)!0.2cm!(\tikztostart) $)!-0.45cm!90:(\tikztostart) $)
	},
	sentence/.style = {
		ampersand replacement = \&,
		column sep = .2cm
	},
	every node/.style = {
		align = center
	}
}

\tikzset{
  >=stealth',
  punkt/.style={
    rectangle,
    rounded corners,
    draw=black, very thick,
    text width=6.5em,
    minimum height=2em,
    text centered},
  pil/.style={
    ->,
    thick,
    shorten <=2pt,
    shorten >=2pt,}
}

\aclfinalcopy 

\title{Data Augmentation via Dependency Tree Morphing for Low-Resource Languages}

\author{Gözde Gül Şahin \\
  UKP Lab, Department of Computer Science \\
  Technische Universität Darmstadt \\
  Darmstadt, Germany \\
  {\tt \small sahin@ukp.informatik.tu-darmstadt.de} \\\And
  Mark Steedman \\
  School of Informatics \\
  University of Edinburgh  \\
  Edinburgh, Scotland \\
  {\tt \small steedman@inf.ed.ac.uk} \\}

\date{}

\begin{document}
\maketitle
\begin{abstract}

Neural NLP systems achieve high scores in the presence of sizable training dataset. Lack of such datasets leads to poor system performances in the case low-resource languages. We present two simple text augmentation techniques using dependency trees, inspired from image processing. We ``crop'' sentences by removing dependency links, and we ``rotate'' sentences by moving the tree fragments around the root. We apply these techniques to augment the training sets of low-resource languages in Universal Dependencies project. We implement a character-level sequence tagging model and evaluate the augmented datasets on part-of-speech tagging task. We show that crop and rotate provides improvements over the models trained with non-augmented data for majority of the languages, especially for languages with rich case marking systems. 

\end{abstract}

\section{Introduction}

 Most recently, various deep learning methods have been proposed for many natural language understanding tasks including sentiment analysis, question answering, dependency parsing and semantic role labeling. Although these methods have reported state-of-the-art results for languages with rich resources, no significant improvement has been announced for low-resource languages. In other words, feature-engineered statistical models still perform better than these neural models for low-resource languages.\footnote{For example, in the case of dependency parsing, recent best results from CoNLL-18 parsing shared task can be compared to the results of traditional language-specific models.} Generally accepted reason for low scores is the size of the training data, i.e., training labels being too sparse to extract meaningful statistics. 
   
 Label-preserving data augmentation techniques are known to help methods generalize better by increasing the variance of the training data. It has been a common practice among researchers in computer vision field to apply data augmentation, e.g., flip, crop, scale and rotate images, for tasks like image classification~\cite{CiresanMS12,KrizhevskySH12}. Similarly, speech recognition systems made use of augmentation techniques like changing the tone and speed of the audio~\cite{KoPPK15,RagniKRG14}, noise addition~\cite{hartmann2016two} and synthetic audio generation~\cite{TakahashiGPG16}. Comparable techniques for data augmentation are less obvious for NLP tasks, due to structural differences among languages. There are only a small number of studies that tackle data augmentation techniques for NLP, such as \citet{ZhangL15} for text classification and \citet{FadaeeBM17a} for machine translation.

 In this work, we focus on languages with small training datasets, that are made available by the Universal Dependency (UD) project. These languages are dominantly from Uralic, Turkic, Slavic and Baltic language families, which are known to have extensive morphological case-marking systems and relatively free word order. With these languages in mind, we propose an easily adaptable, multilingual text augmentation technique based on dependency trees, inspired from two common augmentation methods from image processing: \textbf{cropping} and \textbf{rotating}. As images are cropped to focus on a particular item, we \textbf{crop} the sentences to form other smaller, meaningful and focused sentences. As images are rotated around a center, we \textbf{rotate} the portable tree fragments around the root of the dependency tree to form a synthetic sentence. We augment the training sets of these low-resource languages via crop and rotate operations. In order to measure the impact of augmentation, we implement a unified character-level sequence tagging model. We systematically train separate parts-of-speech tagging models with the original and augmented training sets, and evaluate on the original test set. We show that crop and rotate provide improvements over the non-augmented data for majority of the languages, especially for languages with rich case marking system.

\section{Method}
\label{method}
\begin{figure}
    \centering
      \begin{subfigure}[b]{0.4\textwidth}
      \scalebox{0.8}{
          \begin{tikzpicture}
            \matrix[sentence] {
              \node[word] (1) {Babası}; \&
              \node[word] (2) {ona}; \&
              \node[word] (3) {bir}; \&
              \node[word] (4) {mektup}; \&
              \node[word] (5) {yazdı}; \&
              \node[root] (0) {}; \& 
              \\
              \node[inner ysep=0.3em]{\textit{her father}}; \&
              \node[inner ysep=0.3em]{\textit{to her}}; \&
              \node[inner ysep=0.3em]{\textit{a}}; \&
              \node[inner ysep=0.3em]{\textit{letter}}; \&
              \node[inner ysep=0.3em]{\textit{wrote}}; \&
              \\
            };
            
            \draw[ldep={0.8cm}{det}] (3) to (4);
            \draw[ldep={1.4cm}{nsubj}] (1) to (5);
            \draw[ldep={1.1cm}{iobj}] (2) to (5);
            \draw[ldep={0.8cm}{dobj}] (4) to (5);
            \draw[ldep={0.8cm}{root}] (5) to (0);
            \end{tikzpicture}
            }
      \caption{Dependency analysis}  
      \label{fig:dt}
      \end{subfigure}
     ~
      \begin{subfigure}[b]{0.42\textwidth}
          \includegraphics[width=\textwidth]{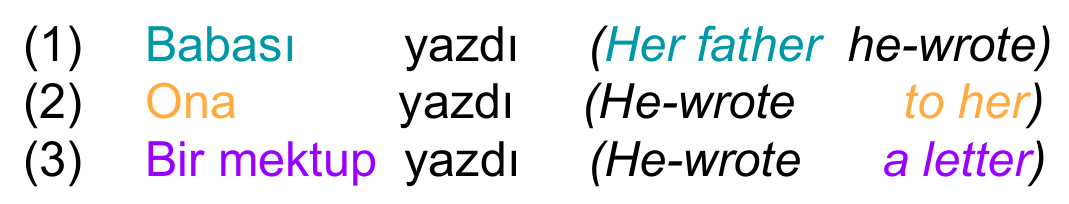}
          \caption{Sentence Cropping}
      \label{fig:crop}
      \end{subfigure}
      ~
      \begin{subfigure}[b]{0.4\textwidth}
          \includegraphics[width=\textwidth]{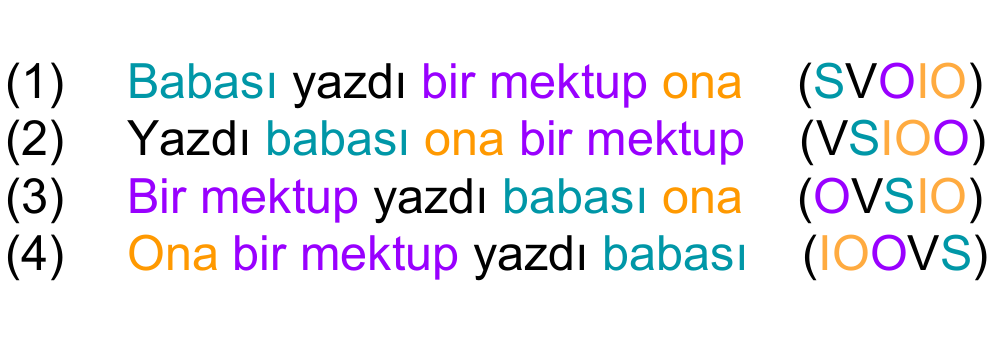}
          \caption{Sentence Rotating}
      \label{fig:flip}
      \end{subfigure}
    \caption{Demonstration of augmentation ideas on the Turkish sentence ``Babası ona bir mektup yazdı'' \textit{(Her father wrote her a letter)}. S: Subject, V: Verb, O:Object, IO: Indirect Object. Arrows are drawn from dependent to head. Both methods are applied to the \textit{Labels of Interest (LOI)}.}
    \label{fig:augment}
   \end{figure}

  We borrow two fundamental label-preserving augmentation ideas from image processing: \textbf{cropping} and \textbf{rotation}. Image cropping can be defined as removal of some of the peripheral areas of an image to focus on the subject/object (e.g., focusing on the flower in a large green field). Following this basic idea, we aim to identify the parts of the sentence that we want to focus and remove the other chunks, i.e., form simpler/smaller meaningful sentences~\footnote{Focus should not be confused with the grammatical category \textsc{FOC}.}. In order to do so, we take advantage of dependency trees which provide us with links to focuses, such as subjects and objects. The idea is demonstrated in Fig.~\ref{fig:crop} on the Turkish sentence given in Fig.~\ref{fig:dt}. Here, given a predicate \textit{(wrote)} that governs a subject \textit{(her father)}, an indirect object \textit{(to her)} and a direct object \textit{(a letter)}; we form three smaller sentences with a focus on the subject (first row in Fig.~\ref{fig:crop}: her father wrote) and the objects (second and third row) by removing all dependency links other than the focus (with its subtree). Obviously, cropping may cause semantic shifts on a sentence-level. However it preserves local syntactic tags and even shallow semantic labels.
       
  Images are rotated around a chosen center with a certain degree to enhance the training data. Similarly, we choose the root as the center of the sentence and rotate the flexible tree fragments around the root for augmentation. Flexible fragments are usually defined by the morphological typology of the language~\cite{wordOrder}. For instance, languages close to analytical typology such as English, rarely have inflectional morphemes. They do not mark the objects/subjects, therefore words have to follow a strict order. For such languages, sentence rotation would mostly introduce noise. On the other hand, large number of languages such as Latin, Greek, Persian, Romanian, Assyrian, Turkish, Finnish and Basque have no strict word order (though there is a preferred order) due to their extensive marking system. Hence, flexible parts are defined as marked fragments which are again, subjects and objects. Rotation is illustrated in Fig.~\ref{fig:flip} on the same sentence.

  In order to investigate the impact of the augmentation, we design a simple sequence tagging model that operates on the character level. Many low-resource languages we deal with in the Experiments section are morphologically rich. Therefore, we use a character-level model to address the rare word problem and to learn morphological regularities among words. 

  For each sentence $s$, we produce a label sequence $\vec{l}$, where $l_t$ refers to POS tag for the $t$-th token. Given $g$ as gold labels and $\theta$ as model parameters we find the values that minimize the negative log likelihood of the sequence:
  \begin{gather} 
    \hat{\theta}=\underset{\theta}{\arg\min} \left( -\sum_{t=1}^n log (p(g_t|\theta,s)) \right)
  \end{gather}
  To calculate $p(l_t|\theta,s)$, we first calculate a word embedding, $\vec{w}$, for each word. We consider words as a sequence of characters ${c_0,c_1,..,c_n}$ and use a bi-LSTM unit to compose the character sequence into $\vec{w}$, as in~\citet{ling2015finding}: 
  \begin{gather} 
    \vec{hw_f}, \vec{hw_b} = \text{bi-LSTM}({c_0,c_1,..,c_n}) \\
    \vec{w} = W_f \cdot \vec{hw_f} + W_b \cdot \vec{hw_b} + b
  \label{eq:comp}
  \end{gather}
  Later, these embeddings are passed onto another bi-LSTM unit:
  \begin{gather} 
    \vec{h_{f}}, \vec{h_{b}} = \text{bi-LSTM}(\vec{w_{t}})
  \end{gather} 
  Hidden states from both directions are concatenated and mapped by a linear layer to the label space. Then label probabilities are calculated by a softmax function: 
  \begin{gather} 
    p(\vec{l_t}|s,p) = \text{softmax}(W_{l}\cdot[\vec{h_{f}};\vec{h_{b}}]+\vec{b_{l}})
  \end{gather}
  Finally the label with the highest probability is assigned to the input. 

\section{Experiments and Results}
  We use the data provided by Universal Dependencies v2.1~\cite{ud21ref} project. Since our focus is on languages with low resources, we only consider the ones that have less than 120K tokens. The languages without standard splits and sizes less than 5K tokens, are ignored. We use the universal POS tags defined by UD v2.1.    

  To keep our approach as language agnostic and simple as possible, we use the following universal dependency labels and their subtypes to extract the focus and the flexible fragment: \textsc{nsubj}~(nominal subject), \textsc{iobj}~(indirect object), \textsc{obj}~(indirect object) and \textsc{obl}~(oblique nominal). These dependency labels are referred to as Label of Interest (LOI). The root/predicate may be a phrase rather than a single token. We use the following relations to identify such cases: \textsc{fixed}, \textsc{flat}, \textsc{cop} (copula) and \textsc{compound}. Other labels such as \textsc{advmod} can also be considered flexible, however are ignored for the sake of simplicity. We enumerate all flexible chunks and calculate all reordering permutations. Keeping the LOI limited is necessary to reduce the number of permutations. We apply reordering only to the first level in the tree. Our method overgeneralizes, to include sequences that are not grammatical in the language in question. We regard the ungrammatical sentences as noise.
  
  Number of possible cropping operations are limited to the number of items that are linked via an LOI to the root. If we call it $n$, then the number of possible rotations would be $(n+1)!$ since $n$ pieces and the root are flexible and can be placed anywhere in the sentence. To limit the number of rotations, we calculate all possible permutations for reordering the sentence and then randomly pick $n$ of them. Each operation is applied with a certain probability $p$ to each sentence,~\textit(e.g., if $p=1$, $n$ number of crops; if $p=0.5$ an average of $n/2$ crops will be done). 

  We use the model in Sec.~\ref{method} to systematically train part-of-speech taggers on original and augmented training data sets. To be able measure the impact of the augmentation, all models are trained with the same hyperparameters. All tokens are lowercased and surrounded with special start-end symbols. Weight parameters are uniformly initialized between $-0.1$ and $+0.1$. We used one layer bi-LSTMs both for character composition and POS tagging with hidden size of 200. Character embedding size is chosen as 200. We used dropout, gradient clipping and early stopping to prevent overfitting for all experiments. Stochastic gradient descent with an initial learning rate as 1 is used as the optimizer. Learning rate is reduced by half if scores on development set do not improve. 

  \begin{table*}
  \centering
    \scalebox{0.85}
    {
    \begin{tabular}{|c|l|l|c|c|c|c|c|c|c|c|}
     \hline
        &   &  &  & \multicolumn{3}{c|}{\textbf{crop}} & \multicolumn{3}{c|}{\textbf{rotate}} &  \\
        \hline
       \textbf{\#Tokens} & \textbf{Lang}  & \textbf{Type} & \textbf{Org} & \textbf{$p=0.3$} & \textbf{$p=0.7$} & \textbf{$p=1$} & \textbf{$p=0.3$} & \textbf{$p=0.7$} & \textbf{$p=1$} & \textit{Imp\%} \\
       \hline
       \multirow{6}{*}{$<20K$} & Lithuanian & \textit{IE, Baltic} & 61.51 & 62.17 & 66.28 & 67.64 & 65.28 & 66.56 & \textbf{68.27} & \textit{10.98} \\
                             & Belarusian & \textit{IE, Slavic} & 83.58 & 83.87 & 85.50 & 85.39 & 84.33 & 85.96 & \textbf{86.11} & \textit{3.03} \\
                             & Tamil      & \textit{Dravidian} & 81.93 & 81.35 & 82.78 & \textbf{84.34} & 83.74 & 83.86 & 83.61 & \textit{2.94} \\                          
                             & Telugu     & \textit{Dravidian} & 90.78 & \textbf{90.85} & 89.88 & 90.50 & 90.36 & 90.29 & 89.95 & \textit{0.07} \\
                             & Coptic     & \textit{Egyptian} & \textbf{95.17} & 94.60 & 94.74 & 94.12 & 95.03 & 94.65 & 94.60 & \textit{-0.15} \\
       \hline
       \multirow{9}{*}{$<80K$} & Irish & \textit{IE, Celtic} & 62.75 & 73.72 & 75.87 & 75.42 & 72.51 & \textbf{76.35} & 76.19 & \textit{21.68} \\
                               & North Sami & \textit{Uralic, Sami} &  86.78 & 86.04 & 87.17 & 87.35 & 87.85 & \textbf{88.04} & 86.65 & \textit{1.45} \\
                               & Hungarian & \textit{Uralic, Ugric} & 85.94 & 86.24 & 86.56 & \textbf{86.62} & 86.49 & 86.37 & 86.60 & \textit{0.80} \\
                               & Vietnamese & \textit{Austro-Asiatic} & 75.16 & \textbf{75.59} & 75.32 & 74.84 & 75.22 & 75.15 & 75.14 & \textit{0.57} \\
                               & Turkish & \textit{Turkic} & 93.49 & 93.53 & 93.56 & 93.89 & 93.60 & 93.82 & \textbf{93.98} & \textit{0.52} \\
                               & Greek & \textit{IE, Greek} & 95.18 & 95.32 & 95.46 & \textbf{95.54} & 95.26 & 95.22 & 95.35 & \textit{0.38} \\
                               & Gothic & \textit{IE, Germanic} & 94.38 & 94.42 & 94.35 & 94.44 & \textbf{94.62} & 94.48 & 94.43 & \textit{0.25} \\
                               & Old Slavic & \textit{IE, Slavic} & 95.36 & 95.34 & 95.33 & \textbf{95.44} & 95.17 & 95.35 & 94.93 & \textit{0.08} \\
                               & Afrikaans & \textit{IE, Germanic} & 94.91 & 94.52 & 94.86 & \textbf{94.93} & 94.73 & 94.70 & 94.92 & \textit{0.0} \\
       \hline
       \multirow{5}{*}{$<120K$} & Latvian & \textit{IE, Baltic} & 91.22 & 91.38 & 91.77 & \textbf{91.78} & 91.69 & 91.62 & 91.76 & \textit{0.61} \\
                                & Danish & \textit{IE, Germanic} & 94.25 & 94.17 & 93.96 & \textbf{94.78} & 94.18 & 94.10 & 94.21 & \textit{0.56} \\
                                & Slovak & \textit{IE, Slavic} & 91.23 & 91.17 & 91.04 & 91.35 & 91.53 & 91.38 & \textbf{91.58} & \textit{0.38} \\
                                & Serbian & \textit{IE, Slavic} & 96.14 & 96.26 & 96.12 & 96.17 & \textbf{96.35} & 96.16 & 96.07 & \textit{0.22} \\
                                & Ukranian & \textit{IE, Slavic} & 94.41 & 94.33 & 94.56 & 94.49 & \textbf{94.57} & 94.38 & 94.47 &  \textit{0.17} \\
       \hline
     \end{tabular}
    }
    \caption{POS tagging accuracies on UDv2.1 test sets. Best scores are shown with \textbf{bold}. Org: Original. $p$: operation probability. \textit{Imp\%}: Improvement over original (Org) by the best model trained with the augmented data.}
  \label{tab:pos_res}
  \end{table*}
  Average of multiple runs for 20 languages are given in Fig.~\ref{tab:pos_res}. Here, Org column refers to our baseline with non-augmented, original training set, where \textit{Imp\%} is the improvement over the baseline by the best crop/flip model for that language. It is evident that, with some minor exceptions, \textit{all} languages have benefited from a type of augmentation. We see that the biggest improvements are achieved on Irish and Lithuanian, the ones with the lowest baseline scores and the smallest training sets~\footnote{Although the total size of the Irish dataset is larger than many, the splits are unbalanced. The training set contains 121 trees while the test has 454 trees.}. Our result on both languages show that both operations reduced the generalization error surprisingly well in the lack of training data. 

\begin{figure*}
  \center
      \includegraphics[width=0.8\textwidth]{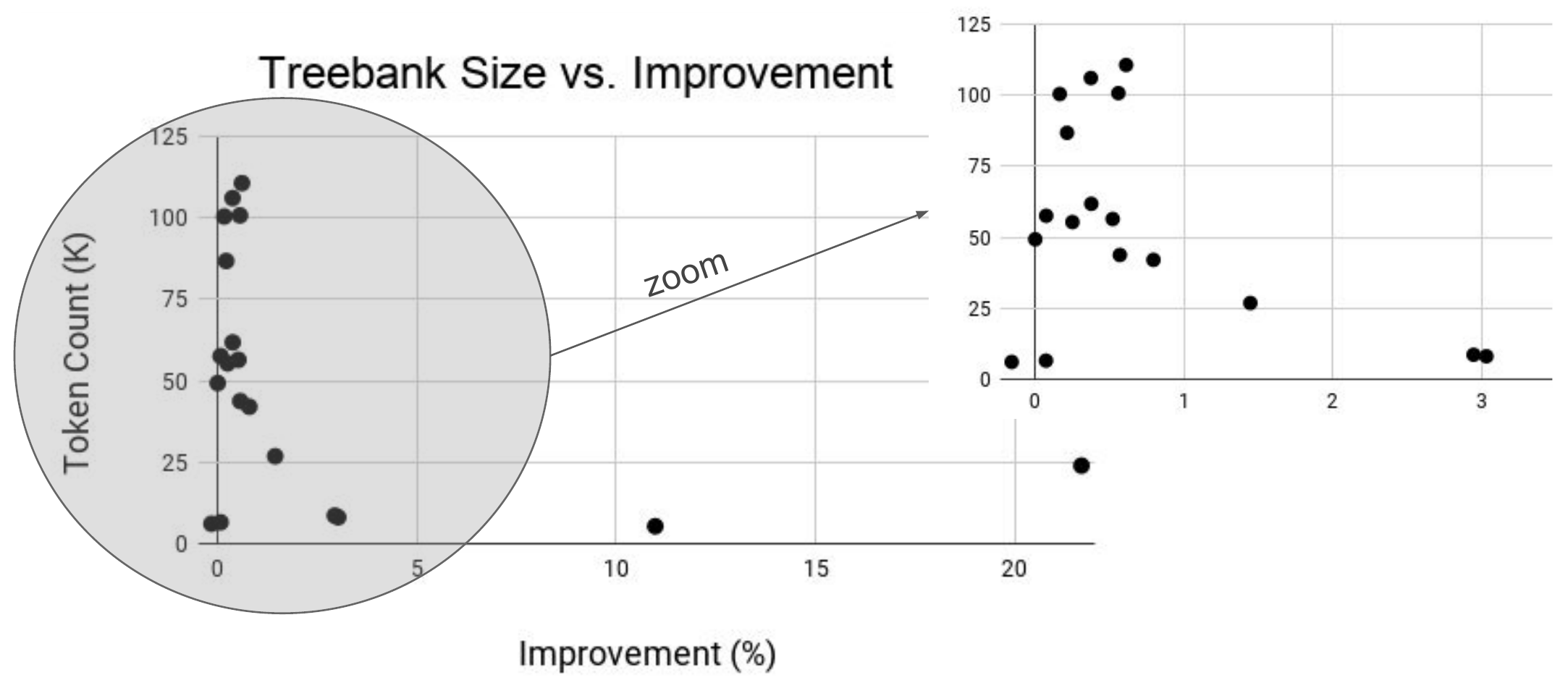}
      \caption{Treebank size versus gain by augmentation}
  \label{fig:exp}
  \end{figure*}

  Tagging results depend on many factors such as the training data size, the source of the treebank (e.g., news may have less objects and subjects compared to a story), and the language typology (e.g., number/type of case markers it uses). In Fig.~\ref{fig:exp}, the relation between the data size and the improvement by the augmentation is shown. Pearson correlation coefficient for two variables is calculated as $-0.35$.
  \paragraph{Indo-European (IE)}: \textbf{Baltic} and \textbf{Slavic} languages are known to have around 7 distinct case markers, which relaxes the word order. As expectedly, both augmentation techniques improve the scores for Baltic (Latvian, Lithuanian) and Slavic (Belarusian, Slovak, Serbian, Ukranian) languages, except for Old Church Slavic (OCS). OCS is solely compiled from bible text which is known to contain longer and passive sentences. We observe that rotation performs slightly better than cropping for Slavic languages. In the presence of a rich marking system, rotation can be considered a better augmenter, since it greatly increases the variance of the training data by shuffling. For \textbf{Germanic} (Gothic, Afrikaans, Danish) languages, we do not observe a repeating gain, due to lack of necessary markers.
  \paragraph{Uralic and Turkic}: Both language types have an extensive marking system. Hence, similar to Slavic languages, both techniques improve the score.
  \paragraph{Dravidian}: Case system of modern Tamil defines 8 distinct markers, which explains the improved accuracies of the augmented models. We would expect a similar result for Telugu. However Telugu treebank is entirely composed of sentences from a grammar book which may not be expressive and diverse.   
\section{Related Work}
   Similar to sentence cropping, \citet{VickreyK08} define transformation rules to simplify sentences (e.g., I was not given a chance to eat - I ate) and shows that enrichening training set with simplified sentences improves the results of semantic role labeling. One of the first studies in text augmentation~\cite{ZhangL15}, replaces a randomly chosen word with its randomly chosen synonym extracted from a thesaurus. They report improved test scores when a large neural model is trained with the augmented set. \citet{robinjia} induce grammar from semantic parsing training data and generate new data points by sampling to feed a sequence to sequence RNN model. \citet{FadaeeBM17a} chooses low-frequency words instead of a random word, and generate synthetic sentence pairs that contain those rare words. 

\section{Discussion}
    Unlike majority of previous NLP augmentation techniques, the proposed methods are meaning-preserving, i.e., they preserve the fundamental meaning of the sentence for most of the tested languages. Therefore can be used for variety of problems such as semantic role labeling, sentiment analysis, text classification. Instead of those problems, we evaluate the idea on the simplest possible task (POS) for the following reasons:
    \begin{itemize}
    \item It gets harder to measure the impact of the idea as the system/task gets complicated due to large number of parameters. 
    \item POS tagging performance is a good indicator of performances of other structured prediction tasks, since POS tags are crucial features for higher-level NLP tasks. 
    \end{itemize}

    Our research interest was to observe which augmentation technique would improve which language, rather than finding one good model. Therefore we have not used development sets to choose one good augmentation model. 

\section{Conclusion and Future Work}
    Neural models have become a standard approach for many NLP problems due to their ability to extract high-level features and generalization capability. Although they have achieved state-of-the-art results in NLP benchmarks with languages with large amount of training data, low-resource languages have not yet benefited from neural models. In this work, we presented two simple text augmentation techniques using dependency trees inspired by image cropping and rotating. We evaluated their impact on parts-of-speech tagging in a number of low-resource languages from various language families. Our results show that:
    \begin{itemize}
      \item Language families with rich case marking systems \textit{(e.g., Baltic, Slavic, Uralic)} benefit both from cropping and rotation. However, for such languages, rotation increases the variance of the data relatively more, leading to slightly better accuracies.
      \item Both techniques provide substantial improvements over the baseline (non-augmented data) when only a tiny training dataset is available. 
    \end{itemize}
    This work aimed to measure the impact of the basic techniques, rather than creating the best text augmentation method. Following these encouraging results, method can be improved by (1) considering the preferred chunk order of the language during rotation, (2) taking language-specific flexibilities into account (e.g., Spanish typically allows free subject inversion (unlike object)). Furthermore, we plan to extend this work by evaluating the augmentation on other NLP benchmarks such as language modeling, dependency parsing and semantic role labeling. The code is available at \url{https://github.com/gozdesahin/crop-rotate-augment}.

\section{Acknowledgements}
Gözde Gül Şahin was a PhD student at Istanbul Technical University and a visiting research student at University of Edinburgh during this study. She was funded by Tübitak (The Scientific and Technological Research Council of Turkey) 2214-A scholarship during her visit to University of Edinburgh. This work was supported by ERC H2020 Advanced Fellowship GA 742137 SEMANTAX and a Google Faculty award to Mark Steedman. We would like to thank Adam Lopez for fruitful discussions, guidance and support during the first author's visit. We thank to the anonymous reviewers for useful comments and to Ilia Kuznetsov for his valuable feedback.

\bibliography{emnlp2018}

\begin{thebibliography}{13}
\expandafter\ifx\csname natexlab\endcsname\relax\def\natexlab#1{#1}\fi

\bibitem[{Ciresan et~al.(2012)Ciresan, Meier, and Schmidhuber}]{CiresanMS12}
Dan~C. Ciresan, Ueli Meier, and J{\"{u}}rgen Schmidhuber. 2012.
\newblock Multi-column deep neural networks for image classification.
\newblock In \emph{2012 {IEEE} Conference on Computer Vision and Pattern
  Recognition, Providence, RI, USA, June 16-21, 2012}, pages 3642--3649.

\bibitem[{Fadaee et~al.(2017)Fadaee, Bisazza, and Monz}]{FadaeeBM17a}
Marzieh Fadaee, Arianna Bisazza, and Christof Monz. 2017.
\newblock Data augmentation for low-resource neural machine translation.
\newblock In \emph{Proceedings of the 55th Annual Meeting of the Association
  for Computational Linguistics, {ACL} 2017, Vancouver, Canada, July 30 -
  August 4, Volume 2: Short Papers}, pages 567--573.

\bibitem[{Futrell et~al.(2015)Futrell, Mahowald, and Gibson}]{wordOrder}
Richard Futrell, Kyle Mahowald, and Edward Gibson. 2015.
\newblock Quantifying word order freedom in dependency corpora.
\newblock In \emph{Proceedings of the Third International Conference on
  Dependency Linguistics (Depling 2015)}, pages 91--100. Uppsala University,
  Uppsala, Sweden.

\bibitem[{Hartmann et~al.(2016)Hartmann, Ng, Hsiao, Tsakalidis, and
  Schwartz}]{hartmann2016two}
William Hartmann, Tim Ng, Roger Hsiao, Stavros Tsakalidis, and Richard~M
  Schwartz. 2016.
\newblock Two-stage data augmentation for low-resourced speech recognition.
\newblock In \emph{Interspeech}, pages 2378--2382.

\bibitem[{Jia and Liang(2016)}]{robinjia}
Robin Jia and Percy Liang. 2016.
\newblock Data recombination for neural semantic parsing.
\newblock In \emph{Proceedings of the 54th Annual Meeting of the Association
  for Computational Linguistics, {ACL} 2016, August 7-12, 2016, Berlin,
  Germany, Volume 1: Long Papers}.

\bibitem[{Ko et~al.(2015)Ko, Peddinti, Povey, and Khudanpur}]{KoPPK15}
Tom Ko, Vijayaditya Peddinti, Daniel Povey, and Sanjeev Khudanpur. 2015.
\newblock Audio augmentation for speech recognition.
\newblock In \emph{{INTERSPEECH} 2015, 16th Annual Conference of the
  International Speech Communication Association, Dresden, Germany, September
  6-10, 2015}, pages 3586--3589.

\bibitem[{Krizhevsky et~al.(2012)Krizhevsky, Sutskever, and
  Hinton}]{KrizhevskySH12}
Alex Krizhevsky, Ilya Sutskever, and Geoffrey~E. Hinton. 2012.
\newblock Imagenet classification with deep convolutional neural networks.
\newblock In \emph{Advances in Neural Information Processing Systems 25: 26th
  Annual Conference on Neural Information Processing Systems 2012. Proceedings
  of a meeting held December 3-6, 2012, Lake Tahoe, Nevada, United States.},
  pages 1106--1114.

\bibitem[{Ling et~al.(2015)Ling, Luis, Marujo, Astudillo, Amir, Dyer, Black,
  and Trancoso}]{ling2015finding}
Wang Ling, Tiago Luis, Luis Marujo, Ramon~F Astudillo, Silvio Amir, Chris Dyer,
  Alan~W Black, and Isabel Trancoso. 2015.
\newblock {Finding function in form: Compositional character models for open
  vocabulary word representation}.
\newblock In \emph{EMNLP}, pages 1520--1530.

\bibitem[{Nivre et~al.(2017)}]{ud21ref}
Joakim Nivre et~al. 2017.
\newblock Universal dependencies 2.1.
\newblock {LINDAT}/{CLARIN} digital library at the Institute of Formal and
  Applied Linguistics, Faculty of Mathematics and Physics, Charles University.

\bibitem[{Ragni et~al.(2014)Ragni, Knill, Rath, and Gales}]{RagniKRG14}
Anton Ragni, Kate~M. Knill, Shakti~P. Rath, and Mark J.~F. Gales. 2014.
\newblock Data augmentation for low resource languages.
\newblock In \emph{{INTERSPEECH} 2014, 15th Annual Conference of the
  International Speech Communication Association, Singapore, September 14-18,
  2014}, pages 810--814.

\bibitem[{Takahashi et~al.(2016)Takahashi, Gygli, Pfister, and
  Gool}]{TakahashiGPG16}
Naoya Takahashi, Michael Gygli, Beat Pfister, and Luc~Van Gool. 2016.
\newblock Deep convolutional neural networks and data augmentation for acoustic
  event detection.
\newblock In \emph{Interspeech}, pages 2982--2986.

\bibitem[{Vickrey and Koller(2008)}]{VickreyK08}
David Vickrey and Daphne Koller. 2008.
\newblock Sentence simplification for semantic role labeling.
\newblock In \emph{{ACL} 2008, Proceedings of the 46th Annual Meeting of the
  Association for Computational Linguistics, June 15-20, 2008, Columbus, Ohio,
  {USA}}, pages 344--352.

\bibitem[{Zhang et~al.(2015)Zhang, Zhao, and LeCun}]{ZhangL15}
Xiang Zhang, Junbo Zhao, and Yann LeCun. 2015.
\newblock Character-level convolutional networks for text classification.
\newblock In \emph{Advances in Neural Information Processing Systems 28}, pages
  649--657. Curran Associates, Inc.

\end{thebibliography}
\bibliographystyle{acl_natbib_nourl}

\end{document}